\documentclass{article}
\usepackage{ijcai26}

\usepackage{times}
\usepackage{soul}
\usepackage{url}
\usepackage[hidelinks]{hyperref}
\usepackage[utf8]{inputenc}
\usepackage[small]{caption}
\usepackage{graphicx}
\usepackage{amsmath}
\usepackage{amssymb}
\usepackage{amsthm}
\usepackage{booktabs}
\usepackage{algorithm}
\usepackage{algorithmic}
\usepackage[switch]{lineno}
\usepackage[nolist,nohyperlinks]{acronym}
\usepackage[shortcuts]{extdash}

\title{KANLab: TensorFlow Playground for Kolmogorov-Arnold Networks}

\author{
    Author Name
    \affiliations
    Affiliation
    \emails
    email@example.com
}

\author{
Lukas Niehaus$^1$$^2$
\and
Robin Rawiel$^1$\and
Ulf Krumnack$^{1}$
\affiliations
$^1$University Osnabrueck\\
$^2$University of Applied Sciences Osnabrueck\\
\emails
\{lukas.niehaus, robin.rawiel, ulf.krumnack\}@uos.de,
}

\begin{acronym}
    \acro{ai}[AI]{Artificial Intelligence}
    \acro{kan}[KAN]{Kolmogorov-Arnold Network}
    \acroplural{kan}[KANs]{Kolmogorov-Arnold Networks}
    \acro{kat}[KAT]{Kolmogorov-Arnold Theorem}
    \acro{gan}[GAN]{Generative Adversarial Network}
    \acroplural{gan}[GANs]{Generative Adversarial Networks}
    \acro{ml}[ML]{Machine Learning}
    \acro{dl}[DL]{Deep Learning}
    \acro{mlp}[MLP]{Multi-layer Perceptron}
    \acroplural{mlp}[MLPs]{Multi-layer Perceptrons}
\end{acronym}

\begin{document}

\maketitle

\begin{abstract}
    KANLab is a browser-based visualization tool designed to interactively teach the core concepts of Kolmogorov-Arnold Networks (KAN), forked from the widely successful TensorFlow Playground. 
    To maximize educational clarity, KANLab omits residual connections and learnable scaling parameters found in modern KAN implementations, ensuring that displayed B-spline activation curves directly correspond to their underlying trainable control points.
    Users can inspect and modify activation functions, initialization schemes and observe gradient descent correcting manual disruptions.
    Furthermore, user can explore phenomena such as activation variance, interactive pruning, and re\-/training within a compact 2D machine learning setting. 
    KANLab is open-source and available at \url{https://kanlab.app/}.
\end{abstract}


\begin{figure*}[t!b]
  \centering
  \includegraphics[width=\textwidth]{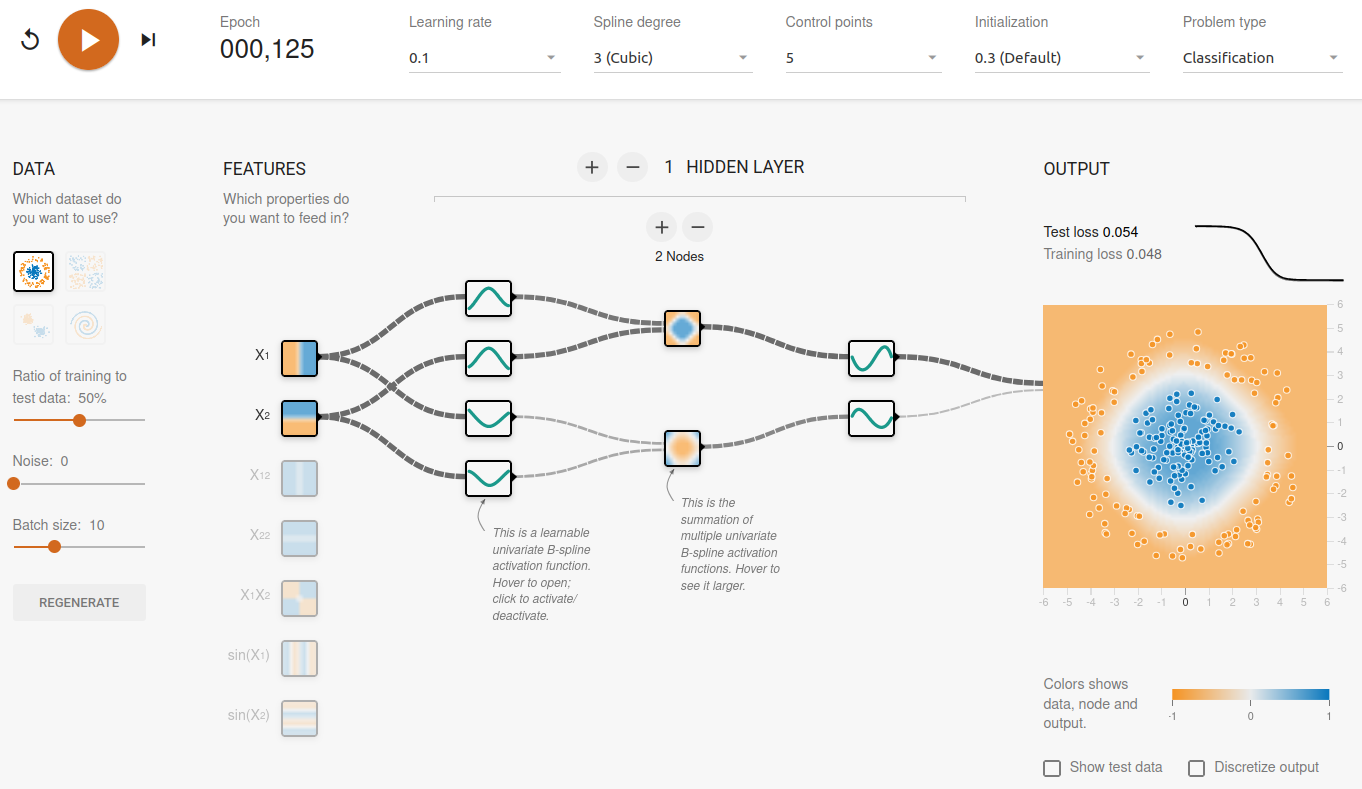}

  \caption{Main view of KANLab, with a two layer \ac{kan} that has 2 input features ($P$), 2 hidden nodes ($Q$) and 1 output Node, represented by squares.
  In between these we visualize the green B-spline activation functions with rectangles.
  Left and right of the rectangles gray dashed lines represent the variance of the incoming (left) and outgoing (right) activation variance by contrasting levels of thickness.}

  
  \label{fig:kanlab}
\end{figure*}

\section{Introduction}

Combining lecture-like instructions with practical application has been shown to enhance the understanding of a variety of topics, including \ac{ai} \cite{turgunbaev_enhancing_2025}. 
A prime example of this is the popular TensorFlow Playground \cite{smilkov_direct-manipulation_2017}, which helps build intuition about the concepts and hyperparameters of \acp{mlp} by letting the user train interactively in the browser.
In addition to being widely successful, it also inspired a whole branch of practice-oriented applications to improve the understanding in different fields of \ac{ai}.
One showcase for this is the Adversarial Playground \cite{norton_adversarial-playground_2017}, which focuses on education about adversarial attacks on already trained neural networks.
Another browser-based rendition, GANLab \cite{kahng_gan_2019} allows the user to train a generator and a discriminator of a \ac{gan} on 2D distributions, while being able to observe their interplay directly.
More recently, the Quantum Machine Learning Playground \cite{debus_quantum_2024} has been released, also leveraging established interactivity and visualization methods.

A relatively new field in \ac{ai} are \acp{kan}, which are popularized in their current form by \cite{liu_kan_2025}.
They have been proposed as an interesting new tool in the repertoire of a deep learning researcher and can be used as an alternative to classical \acp{mlp}, promising more compact architecture due to more powerful activation functions, resulting in more interpretable models.
\acp{mlp} layers sum up weighted inputs and then pass them through a static non-linear activation function (like ReLU) to solve non-linear problems.
\acp{kan} layers, on the other hand, use trainable non-linear activation functions and sum up their results on nodes.
While the primary success of \acp{kan} is in symbolic formula representation tasks \cite{yu_kan_2024}, they are also frequently experimented with for classification tasks \cite{yang_medkan_2025,hadj_kilani_kolmogorov-arnold_2024,jamali_how_2024,ibrahum_how_2024}.

Regarding \ac{kan}-specific projects, the KAN-Playground \cite{solovyeva_qksolovkan-playground_2025} demonstrates a feature-rich Gradio implementation.
While the project is readily accessible via Hugging Face Spaces, training is still computed server-side. While this enables a close-to feature-complete and high-performant experience, this API-based approach comes at the drawback of offering limited responsiveness and interactivity.
For example, the architecture is represented via an image, that is updated once per epoch.
Furthermore, modern \ac{kan} implementations add residual connections and learnable scaling/shifting parameters on top of the underlying univariate function. 
While these additions are important for performance, they distort the relationship between the activation visualization and the underlying spline parameterization, making the visualizations harder to interpret from an educational perspective.

Due to the close similarities between \acp{kan} and \acp{mlp}, and the established educational design decisions of the TensorFlow Playground, we created KANLab\footnote{\url{https://kanlab.app/}} as a fork of it.
This allows us to create more specialized visualizations and even enables users to drag individual control points during training and observe how gradient descent corrects these 
manual disruptions in real time.
KANLab aims to fill the gap between people who are new to \acp{kan} and want to understand the core concepts, and \ac{kan}-experts who want to communicate or probe KAN 
behavior in an accessible environment.

\section{Kolmogorov-Arnold Networks}
Following their introduction by \cite{liu_kan_2025}, \acp{kan} have garnered attention as a potential alternative to \acp{mlp}. 
However, learning their concepts through practical experience is still challenging.
KANLab addresses this gap by providing a browser interface for interactive design, training, and analysis of \acp{kan} in real time. 
Figure~\ref{fig:kanlab} shows a screenshot of the main view, which dynamically visualizes the network architecture.
It also allows users to modify hyperparameters, select input features, and observe immediate effects on the model's behavior.  
The theoretical foundation of KANLab is the Kolmogorov-Arnold Representation Theorem, which states that any multivariate continuous function on a bounded domain can be represented as a finite composition of univariate functions and addition:
\begin{equation}
f(x_1,\dots,x_n) = 
    \sum_{q=1}^{Q} 
        \Phi_q
        \left(  \sum_{p=1}^P\phi_{q,p}(x_p) \right),
\end{equation} where $\Phi_q,\phi_{q,p}:\mathbb{R}\to\mathbb{R}$ and $Q=2n+1, P=n$.  
The theorem guarantees a two-layer solution with $Q=2n+1$ nodes, but practical implementations often differ form by adjustments in width and also in depth.
Figure~\ref{fig:kanlab} shows an example with $P=2$ input features ($X_1$, $X_2$) and $Q=2$ hidden nodes. 
Unlike \acp{mlp} with fixed activation functions, \acp{kan} use B-splines that adapt during training, allowing for more flexible function approximation.
The main view of KANLab shows the network architecture and animated B-splines illustrate information flow in real time. 
The network is trained using batched stochastic gradient descent minimizing the mean squared error loss.
During this training the user can observe the evolving B-splines to gain insights on how \acp{kan} are able to decompose complex problems into simpler and trainable components.

\section{Activation Functions}

Extensive effort has gone into utilizing different basis functions to form the univariate activation functions.  
They include B-splines, Chebyshev and Jacobi Polynomials, ReLU compositions, Fourier series, Gaussian kernels, wavelets, finite basis partitions and sinc functions \cite{noorizadegan_practitioners_2026}.
Among these bases the most widely adopted one is the B-spline, which was also introduced in the \ac{kan} formulation by \cite{liu_kan_2025}.
KANLab uses this B-spline-based approach, since their theory allows for some of the strongest guarantees and has the firmest theoretical footing among \ac{kan} bases \cite{noorizadegan_practitioners_2026}.
The B-spline $\varphi$ is defined as: 
\begin{equation}
    \varphi(x)=\sum_{n=0}^{N-1}c_{n}\,B_{nk}(x)
\end{equation}
where $B_{nk}(x)$ are fixed B-spline basis functions of degree $k$ (computed using de Boor’s algorithm), and $c_{n}$ are trainable control points. 
The number of control points $N$ can be set via a dropdown menu in the top bar (Figure~\ref{fig:kanlab}).

\begin{figure}[ht]
  \centering
  \includegraphics[width=\columnwidth]{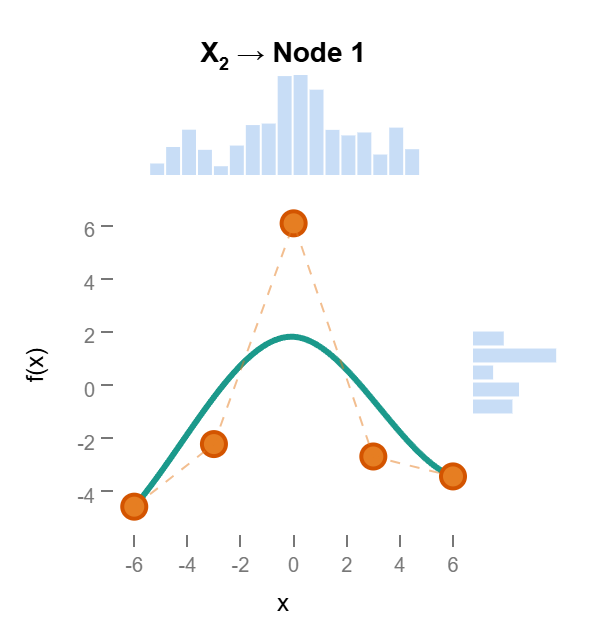}
  \caption{
  Visualization of an activation function parameterized by a B-spline and its effect on activations. The green curve shows the current B-spline activation, while the orange points denote its learnable control points. The top histogram depicts the distribution of incoming activations, and the right histogram shows the distribution of outgoing activations after applying the activation function. Adjusting the orange control points changes the shape of the B-spline and, consequently, the mapping from inputs to outputs.}
  \label{fig:splinechart}
\end{figure}

Figure \ref{fig:splinechart} visualizes the $N=5$ control points as orange circles. 
In KANLab the view depicted in Figure~\ref{fig:splinechart} is displayed when hovering over the rectangle of an activation function from Figure~\ref{fig:kanlab}. 
The TensorFlow Playground and KANLab both generate their datasets in the interval of $[-6,6]$.
To make sure that all generated activations are represented on the B-splines, the control points are uniformly distributed on the same interval.
The control points can be changed by dragging and dropping them to new positions.
The effect of control point manipulation is directly observable in the decision boundaries of the final output and in the intermediate nodes.
The incoming activation for a B-spline in the hidden layer is computed by summing over outputs of the B-splines from the preceding layer.
Since the trainable control points are not bounded, the outgoing activations of a B-spline can exceed the control point domain $[-6,6]$ of the next B-spline.
KANLab handles these cases by clipping the incoming activations of a B-spline into that interval.
\\

\section{Initialization}
The topic of KAN initialization is still in development and KANLab supports multiple initialization methods. 
Each method determines the initialization variance of the control points. \cite{liu_kan_2025} recommend initializing the control points with small values.
KANLab offers preset variances [0.1, 0.3, 0.5, 1.0, 5.0] sampled from a uniform distribution. 
KANLab also supports the LeCun- and Glorot-inspired, variance preserving initialization schemes proposed by \cite{rigas_initialization_2025}.
Additionally, KANLab supports identity as initialization method, which essentially starts the network with linear pass-through transformations by either initializing the control points with an identity function or its inverse.
The variance is selectable via the top bar of Figure \ref{fig:kanlab}.
Users can experiment and observe that small-variance initialization leads to near-zero activations initially. 
Identity initialization, on the other hand, preserves input structure, making it easier to see how training reshapes the activation functions.

\section{Use Cases}
Just like for the TensorFlow Playground, the experimental setup is serialized in the URL's GET parameters.
This allows a teacher to save the URL of an interesting setup and share it with students.
Example setups could show the differences in training behavior between wide and deep \acp{kan}.
The feature of manually adjusting the control points for each spline can also be used to train a \ac{kan} by hand.
To do so, the user hovers over an activation function in Figure~\ref{fig:kanlab}, so that a small window appears, which looks like Figure~\ref{fig:splinechart}.
There, the user can drag and drop the individual control points until the desired result in the following node is reached.
In Figure~\ref{fig:kanlab} we see thick dark lines on the upper path and thinner lines on the lower path. This means that the activation variance is higher on the upper path which also indicates a stronger information flow.
Each activation function on the edges and every (non-output) summation node in Figure \ref{fig:kanlab} can be activated or deactivated individually during training by clicking on them (long click for touchscreens). 
When a spline in the upper path in Figure~\ref{fig:kanlab} gets pruned the loss increases and in the output we see that the classification error grows.
The pruned node can now no longer pass information to the next layer.
If training continues, the activation variance on the lower path increases and the classification error decreases, demonstrating that the network now routes information through the remaining connections.

\section{Conclusion}

We present KANLab, an interactive browser application, offering a familiar TensorFlow Playground-based interface.
Our deliberate design choices favor educational benefits and clarity over performance, in order to close the gap between \ac{kan}-beginners and \ac{kan}-experts.
Future work includes the addition of regularization methods.
\vspace{2cm}

\appendix

\bibliographystyle{named}
\bibliography{KANLab}

@inproceedings{rigas_initialization_2025,
title={{Initialization Schemes for Kolmogorov{\textendash}Arnold Networks: An Empirical Study}},
author={Spyros Rigas and Dhruv Verma and Georgios Alexandridis and Yixuan Wang},
booktitle={The Fourteenth International Conference on Learning Representations},
year={2026},
url={https://openreview.net/forum?id=dwNXKkiP51}
}

@misc{solovyeva_qksolovkan-playground_2025,
	title = {qksolov/kan-playground},
	url = {https://github.com/qksolov/kan-playground},
	urldate = {2026-02-04},
	author = {Solovyeva, Ekaterina},
	month = dec,
	year = {2025},
	note = {original-date: 2025-02-25T17:43:08Z},
}

@inproceedings{liu_kan_2025,
  author = {Ziming Liu and Yixuan Wang and Sachin Vaidya and Fabian Ruehle and James Halverson and Marin Soljacic and Thomas Y. Hou and Max Tegmark},
  title = {{KAN}: {K}olmogorov{\textendash}{A}rnold {N}etworks},
  booktitle = {The Thirteenth International Conference on Learning Representations},
  year = {2025},
  url = {https://openreview.net/forum?id=Ozo7qJ5vZi},
}

@misc{smilkov_direct-manipulation_2017,
	title = {Direct-{Manipulation} {Visualization} of {Deep} {Networks}},
	url = {http://arxiv.org/abs/1708.03788},
	doi = {10.48550/arXiv.1708.03788},
	urldate = {2026-02-05},
	publisher = {arXiv},
	author = {Smilkov, Daniel and Carter, Shan and Sculley, D. and Viégas, Fernanda B. and Wattenberg, Martin},
	month = aug,
	year = {2017},
	note = {arXiv:1708.03788 [cs]},
}

@article{debus_quantum_2024,
	title = {Quantum {Machine} {Learning} {Playground}},
	volume = {44},
	issn = {0272-1716, 1558-1756},
	url = {http://arxiv.org/abs/2507.17931},
	doi = {10.1109/MCG.2024.3456288},
	number = {5},
	urldate = {2026-02-05},
	journal = {IEEE Computer Graphics and Applications},
	author = {Debus, Pascal and Issel, Sebastian and Tscharke, Kilian},
	month = sep,
	year = {2024},
	note = {arXiv:2507.17931 [quant-ph]},
	pages = {40--53},
}

@inproceedings{norton_adversarial-playground_2017,
  title={{Adversarial-Playground: A visualization suite showing how adversarial examples fool deep learning}},
  author={Norton, Andrew P and Qi, Yanjun},
  booktitle={Visualization for Cyber Security (VizSec), 2017 IEEE Symposium on},
  pages={1--4},
  year={2017},
  organization={IEEE}
}

@article{turgunbaev_enhancing_2025,
	title = {Enhancing student understanding of artificial intelligence through practical neural network applications},
	volume = {1},
	copyright = {Copyright (c) 2025},
	issn = {3041-7848},
	url = {https://integrumpublication.org/index.php/ajste/article/view/67},
	language = {en},
	number = {6},
	urldate = {2026-02-05},
	journal = {Academic Journal of Science, Technology and Education},
	author = {Turgunbaev, Rashid},
	month = oct,
	year = {2025},
	pages = {36--40},
}

@misc{yu_kan_2024,
	title = {{KAN} or {MLP}: {A} {Fairer} {Comparison}},
	shorttitle = {{KAN} or {MLP}},
	url = {http://arxiv.org/abs/2407.16674},
	doi = {10.48550/arXiv.2407.16674},
	urldate = {2026-02-12},
	publisher = {arXiv},
	author = {Yu, Runpeng and Yu, Weihao and Wang, Xinchao},
	month = aug,
	year = {2024},
	note = {arXiv:2407.16674 [cs]},
}

@article{kahng_gan_2019,
	title = {{GAN} {Lab}: {Understanding} {Complex} {Deep} {Generative} {Models} using {Interactive} {Visual} {Experimentation}},
	volume = {25},
	issn = {1941-0506},
	shorttitle = {{GAN} {Lab}},
	url = {https://ieeexplore.ieee.org/abstract/document/8440049},
	doi = {10.1109/TVCG.2018.2864500},
	number = {1},
	urldate = {2026-02-09},
	journal = {IEEE Transactions on Visualization and Computer Graphics},
	author = {Kahng, Minsuk and Thorat, Nikhil and Chau, Duen Horng and Viégas, Fernanda B. and Wattenberg, Martin},
	month = jan,
	year = {2019},
	pages = {310--320},
}

@misc{yang_medkan_2025,
	title = {{MedKAN}: {An} {Advanced} {Kolmogorov}-{Arnold} {Network} for {Medical} {Image} {Classification}},
	shorttitle = {{MedKAN}},
	url = {http://arxiv.org/abs/2502.18416},
	doi = {10.48550/arXiv.2502.18416},
	urldate = {2026-02-05},
	publisher = {arXiv},
	author = {Yang, Zhuoqin and Zhang, Jiansong and Luo, Xiaoling and Lu, Zheng and Shen, Linlin},
	month = feb,
	year = {2025},
	note = {arXiv:2502.18416 [cs]},
}

@article{hadj_kilani_kolmogorov-arnold_2024,
	title = {Kolmogorov-{Arnold} {Networks}: {Key} {Developments} and {Uses}},
	shorttitle = {Kolmogorov-{Arnold} {Networks}},
	url = {https://hal.science/hal-04663521},
	doi = {10.32388/7nncaa},
	urldate = {2026-02-05},
	journal = {Qeios},
	author = {Hadj Kilani, Bochra},
	month = jun,
	year = {2024},
}

@misc{noorizadegan_practitioners_2026,
	title = {A {Practitioner}'s {Guide} to {Kolmogorov}-{Arnold} {Networks}},
	url = {http://arxiv.org/abs/2510.25781},
	doi = {10.48550/arXiv.2510.25781},
	urldate = {2026-02-05},
	publisher = {arXiv},
	author = {Noorizadegan, Amir and Wang, Sifan and Ling, Leevan and Dominguez-Morales, Juan P.},
	month = jan,
	year = {2026},
	note = {arXiv:2510.25781 [cs]},
}

@article{ibrahum_how_2024,
	title = {How {Resilient} {Are} {Kolmogorov}–{Arnold} {Networks} in {Classification} {Tasks}? {A} {Robustness} {Investigation}},
	volume = {14},
	copyright = {http://creativecommons.org/licenses/by/3.0/},
	issn = {2076-3417},
	shorttitle = {How {Resilient} {Are} {Kolmogorov}–{Arnold} {Networks} in {Classification} {Tasks}?},
	url = {https://www.mdpi.com/2076-3417/14/22/10173},
	doi = {10.3390/app142210173},
	language = {en},
	number = {22},
	urldate = {2026-02-05},
	journal = {Applied Sciences},
	author = {Ibrahum, Ahmed Dawod Mohammed and Shang, Zhengyu and Hong, Jang-Eui},
	month = nov,
	year = {2024},
}

@article{jamali_how_2024,
	title = {How to {Learn} {More}? {Exploring} {Kolmogorov}–{Arnold} {Networks} for {Hyperspectral} {Image} {Classification}},
	volume = {16},
	copyright = {http://creativecommons.org/licenses/by/3.0/},
	issn = {2072-4292},
	shorttitle = {How to {Learn} {More}?},
	url = {https://www.mdpi.com/2072-4292/16/21/4015},
	doi = {10.3390/rs16214015},
	language = {en},
	number = {21},
	urldate = {2026-02-05},
	journal = {Remote Sensing},
	author = {Jamali, Ali and Roy, Swalpa Kumar and Hong, Danfeng and Lu, Bing and Ghamisi, Pedram},
	month = oct,
	year = {2024},
}

\end{document}